\documentclass[runningheads]{llncs}
\usepackage{graphicx}

\usepackage{tikz}
\usepackage{subcaption}
\usepackage{comment}
\usepackage{amsmath,amssymb} 
\usepackage{color}
\usepackage{hyperref}

\usepackage[accsupp]{axessibility}  

\usepackage{lmodern}
\usepackage{booktabs}
\usepackage{float}

\usepackage[width=122mm,left=12mm,paperwidth=146mm,height=193mm,top=12mm,paperheight=217mm]{geometry}

\begin{document}
\pagestyle{headings}
\mainmatter
\def\ECCVSubNumber{10112}  

\title{Synthetica: Large Scale Synthetic Data Generation for Robot Perception} 

\authorrunning{R. Singh et al.} 
\author{Ritvik Singh\inst{1,2}, Jingzhou Liu\inst{1,2}, Karl Van Wyk\inst{1}, Yu-Wei Chao\inst{1}, Jean-Francois Lafleche\inst{1}, Florian Shkurti\inst{2}, Nathan Ratliff\inst{1}, Ankur Handa\inst{1}}
\institute{NVIDIA \and University of Toronto}

\maketitle

\begin{abstract}

Vision-based object detectors are a crucial basis for robotics applications as they provide valuable information about object localisation in the environment. These need to ensure high reliability in different lighting conditions, occlusions, and visual artifacts, all while running in real-time. Collecting and annotating real-world data for these networks is prohibitively time consuming and costly, especially for custom assets, such as industrial objects, making it untenable for generalization to in-the-wild scenarios. To this end, we present Synthetica, a method for large-scale synthetic data generation for training robust state estimators. This paper focuses on the task of object detection, an important problem which can serve as the front-end for most state estimation problems, such as pose estimation. Leveraging data from a photorealistic ray-tracing renderer, we scale up data generation, generating 2.7 million images, to train highly accurate real-time detection transformers. We present a collection of rendering randomization and training-time data augmentation techniques conducive to robust sim-to-real performance for vision tasks. We demonstrate state-of-the-art performance on the task of object detection while having detectors that run at 50--100Hz which is 9 times faster than the prior SOTA \footnote{Official results on the BOP leaderboard: \url{https://bop.felk.cvut.cz/leaderboards/detection-bop22/ycb-v/}}. We further demonstrate the usefulness of our training methodology for robotics applications by showcasing a pipeline for use in the real world with custom objects for which there do not exist prior datasets. Our work highlights the importance of scaling synthetic data generation for robust sim-to-real transfer while achieving the fastest real-time inference speeds. Videos and supplementary information can be found at \url{https://sites.google.com/view/synthetica-vision}.
\end{abstract}

\section{Introduction}
Object detection is the problem of localising objects of interest in the image with a 2D bounding box and classifying them. It is an important problem in robotics as it serves as the first step in most pose estimation methods. State estimation for robotics tasks requires real-time speeds to be used in downstream control policies while also being robust to motion blur, occlusion, varying lighting conditions, and other imaging artifacts. There has been significant progress in the task of instance-level object detection~\cite{labbé2020cosypose,wang2021gdrnet,hu2022perspective}. These networks train on relatively small datasets, such as the original YCB-Video (YCBV) and T-Less (on the order of ~200K images), consisting of both data collected from the real world and simulation and display impressive performances on the test sets. 

We improve over prior methods in 2 key areas: data collection and inference speed. First, there is an over-reliance on data collected in the real world. Performing data collection and annotation on real-world images is exceedingly time-consuming and expensive, not to mention its difficultly in scaling. For example, in an industrial setting, it is fair to assume access to CAD models of objects, but it is infeasible to assume people will be able to collect and annotate hundreds of thousands of images of the objects in different configurations. It also becomes difficult to adequately capture the wide-ranging distribution of possible images. In order to address this, we generate 2.7 million images using a photorealistic real-time ray-tracing renderer with extensive rendering randomizations along with training-time data augmentations. This way, we are able to generate a wide variety of configurations to ensure that our networks do not overfit to a narrow training distribution. In addition, most existing works tend to sacrifice inference speed for increased network performance, which is detrimental for robotics tasks, such as manipulation, due to the low-latency requirement of those networks. We optimize our model using various kernel fusion techniques to allow for real-time speeds. 

We show that our method of large-scale synthetic data generation is able to achieve state-of-the-art performance on the YCBV object detection benchmark. Furthermore, using TensorRT, we are able to reduce the latency of our network such that the inference speed is 1.5 times faster than the next fastest method, and with slightly worse performance we can achieve speeds that are almost 7 times faster than the state-of-the-art. Finally, we demonstrate the effectiveness of our approach by developing a simple pipeline for scanning objects in the real world in order to create detectors on custom datasets for user-specific domains.

\section{Related Work}
\subsection{Object Detectors}
Object detection can generally be broken down into category level and instance level. The category level problem involves regressing the bounding box for object categories (e.g. ``dog", ``cat", etc.) whereas the instance level is more concerned with specific instances of categories. Examples of category level datasets include ImageNet~\cite{russakovsky2015imagenet} and COCO~\cite{lin2015microsoft} whereas examples of instance level datasets include the YCBV dataset~\cite{xiang2018posecnn} and TLESS~\cite{hodan2017tless}. Given that most RGB-based state-estimation is generally done at the instance-level (otherwise inferring properties such as the pose of categories becomes an ill-posed problem without assumption of any metric scale), the focus of this paper will be on instance level object detection. For instance level object detection, the BOP leaderboard is dominated by~\cite{wang2021gdrnet,hu2022perspective}, which respectively use YOLOv8~\cite{Ultralytics2023} and FCOS~\cite{tian2019fcos}. They employ single-stage, anchor-free detectors in order to achieve their performance. They obtain their state-of-the-art results by training on both real images and simulated PBR images of the objects. However, they only use 50K synthetic images for the YCBV dataset. The top performing method is only able to run inference at 5Hz while the fastest method on the leaderboard operates at 33Hz. There is a noticeable trend of shifting from ResNet~\cite{he2015deep} backbones to larger ones such as ConvNext~\cite{liu2022convnet} which comes with an increase in mAP scores and a decrease in inference speed.

While most of the impressive works use convolution-based neural networks, the growing popularity of transformers in sequence modelling tasks has led to the rise of transformer-based architectures. Detection transformers~\cite{carion2020endtoend} have recently emerged as promising architectures that can scale to larger dataset sizes. While they also make use of a convolutional backbone, they treat object detection as a set prediction problem, and are able to avoid using hand-designed modules such as non-maximum suppression or anchor generation. RT-DETR~\cite{lv2023detrs} builds upon the detection transformer architecture to make them capable of real-time inference, which is crucial for robotics settings.

The recent rise of Visual Language Models~\cite{radford2021learning,yang2023dawn} has brought about a new type of classification: Open Vocabulary Detection~\cite{wu2024open}. This is the most general type of detection problem which attempts to identify and localise objects in an image that were unseen at training time by making use of natural language text. Some examples include GPT-4V~\cite{yang2023dawn}, OWL-ViT~\cite{minderer2022simple}, Llava~\cite{liu2023llava,liu2023improvedllava}, and GLEE~\cite{wu2023GLEE}. While they show a remarkable ability to generalize to unseen objects in the training distribution, they are fundamentally limited in two ways for robotics use-cases: inference speed and robustness for bespoke objects. These VLMs utilize some form of large transformer-like architecture to encode natural language prompts, and as a result, are unable to run at real-time speeds of more than 30Hz. Furthermore, for relatively rare objects that are not as common in internet-scale datasets, VLMs struggle to robustly detect them in videos. 

\subsection{Scaling Synthetic Data}
With the rising popularity of diffusion models~\cite{song2021scorebased,ho2020denoising}, there has been a lot of work involved in using them for synthetic data generation. Being able to query a generative model to produce images of a text prompt allows for a virtually unlimited amount of synthetic data. This has been used to great success in tasks of image classification~\cite{azizi2023synthetic,he2023synthetic}. There has been an extensive amount of work using them to produce more in-context data augmentations~\cite{trabucco2023effective}, and for self-supervised representation learning methods~\cite{tian2023stablerep}. Lastly, there have been many studies analysing the scaling laws of synthetic data as compared to real data~\cite{fan2023scaling}. With all of these examples, the vision problems do not involve dense annotations. For low-level vision tasks, which are also of interest to roboticists, such as detection, segmentation, optical flow, etc., it is not as straightforward to extract the dense annotations required to supervise a model. Hence, we use Omniverse Isaac Sim~\cite{isaacsim2024}, which provides a real-time ray tracing renderer to synthetically generate realistic images with perfect groundtruth annotations. 

\subsection{Synthetic Data for Pose Estimation}
A lot of the state estimation that is used for downstream robotics tasks comes in the form of estimating the pose of objects in the scene~\cite{handa2024dextreme,openai2019learning,wen2022catgrasp,wen2022demonstrate}. For example, DeXtreme~\cite{handa2024dextreme} and Learning Dexterity~\cite{openai2019learning} generate millions of synthetic images to train their pose estimators. However, their work is limited to only generating images of a cube with very discernible textures to aid in the pose estimation process. Other works such as ReorientBot~\cite{wada2022reorientbot} do pose estimation on YCB objects to facilitate robust object re-posing. Existing pose estimation methods such as ~\cite{wang2021gdrnet,hu2022perspective} train on 50k images generated with blender proc~\cite{Denninger2023} alongside the real images for the YCBV dataset. Others such as ~\cite{labbé2020cosypose,labbé2022megapose,wen2023foundationpose} generate several million images to train their pose estimators. They all rely on using off-the-shelf detectors to provide image crops of objects of interest in order to regress their pose. Moreover, it has been noticed that false or missing detections is what frequently bottlenecks the pose estimation performance~\cite{wen2023foundationpose}. While the aforementioned work does scale synthetic data for object pose estimation, there has been limited work in applying it to object detectors. Additionally, ~\cite{hodan2019photorealistic} generate 600k images using photorealistic renderers to generate a large synthetic dataset for various vision tasks. However, their method revolves around randomizing object configurations in only 6 scenes, thus severely limiting the diversity of their data.

\section{Approach}\label{sec:approach}


\subsection{Synthetic Data Generation}
\begin{figure*}[h]
\centering
    \includegraphics[width=0.32\textwidth]{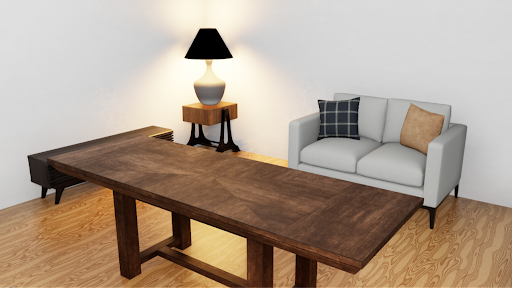}
    \hfill
    \includegraphics[width=0.32\textwidth]{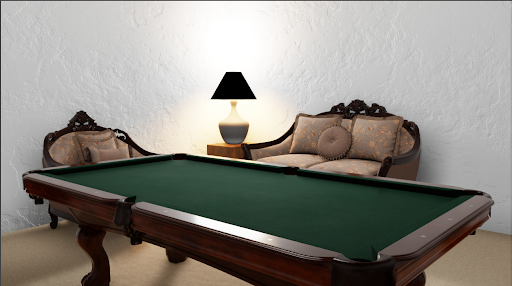}
    \hfill
    \includegraphics[width=0.32\textwidth]{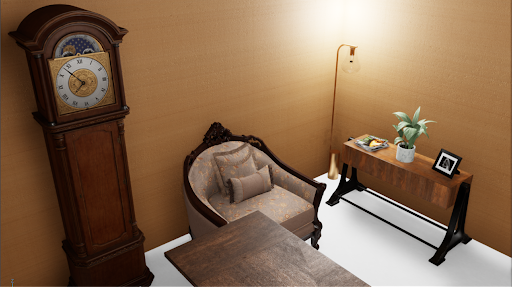}
\caption{Examples of procedurally generated scenes.}
\label{fig:empty_scenes}
\end{figure*}

\begin{figure*}[ht]
    \includegraphics[width=0.45\textwidth]{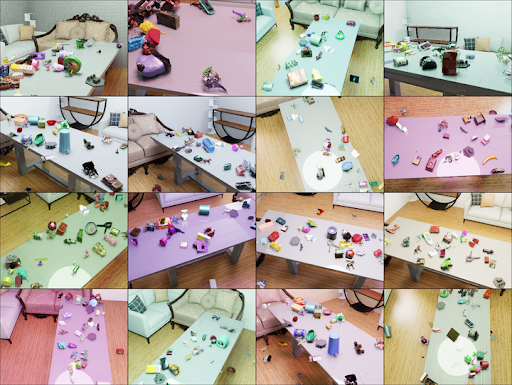}
    \hfill
    \includegraphics[width=0.45\textwidth]{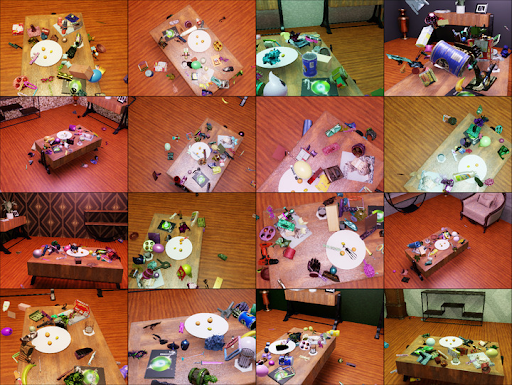}
\caption{Examples of rendering randomizations applied to individual scenes.}
\label{fig:objects_scenes}
\end{figure*}

\begin{figure*}[h]
\centering
    \includegraphics[width=0.32\textwidth]{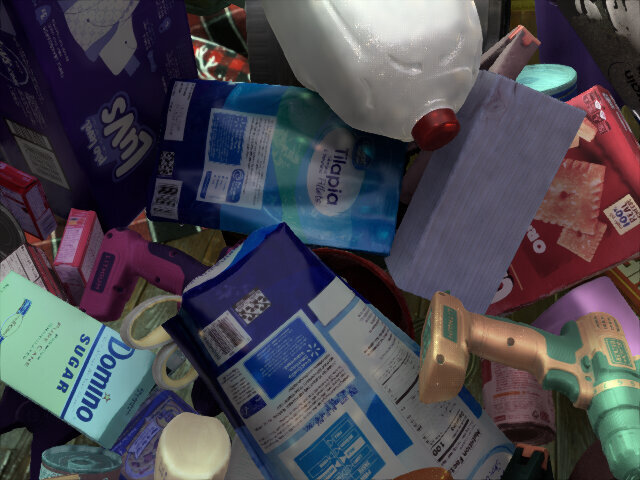}
    \hfill
    \includegraphics[width=0.32\textwidth]{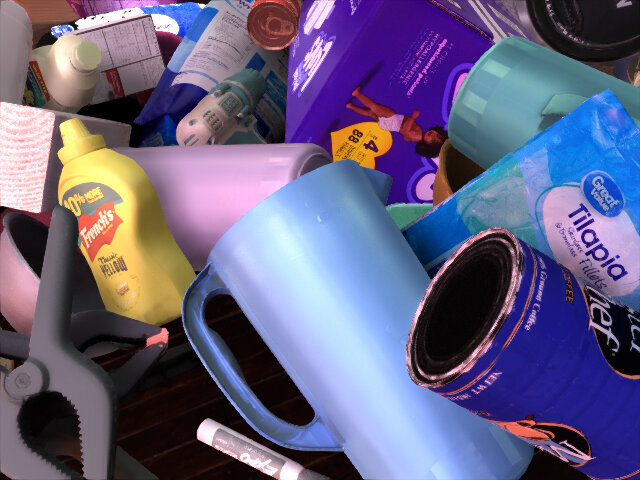}
    \hfill
    \includegraphics[width=0.32\textwidth]{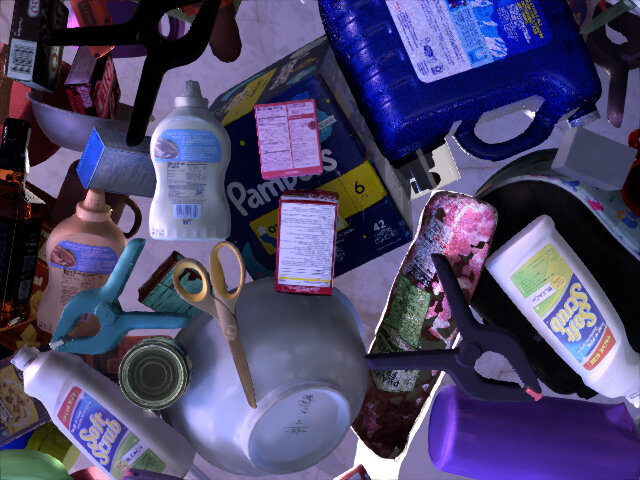}
\caption{Examples of images generated by dropping objects onto an invisible plane.}
\label{fig:hdri_scenes}
\end{figure*}

We use NVIDIA Omniverse Isaac Sim with Replicator~\cite{isaacsim2024} to generate 2.7M images in simulation. We render data by either procedurally generating an indoor room and dropping objects in it, or dropping objects with a random HDRI background. In order to avoid false-positive detections in the real world, we not only dropped YCB objects, but we also collected other \emph{distractor} objects from Objaverse~\cite{deitke2022objaverse}, Google Scanned Objects~\cite{downs2022google}, and the NVIDIA asset library to drop with the YCB objects. This added more variety in the scene and also increased the amount of occlusions on the YCB objects. To procedurally generate an indoor room, we set the room width to be a random number between 4.5 and 5, and set the length to be a random scaling factor between 1.0 and 1.1 times the width. Using the NVIDIA asset library in Isaac Sim, we place random furniture around the perimeter of the room and place a table in the middle. Examples of these are shown in Figure~\ref{fig:empty_scenes}. From there, we generate images by dropping objects within the boundaries of this table and had the camera move around the room while looking at the center of the table (example images shown in Figure~\ref{fig:objects_scenes}). Every 3000 frames, we would use a new scene. In the case of HDRI background randomization, we dropped objects from a height ranging from 1 meter to 5 meters in a 1x1 meter enclosure with invisible walls to funnel the objects onto an invisible ground plane. We would have a camera randomly move around the clutter and periodically change the HDRI background. Example images are shown in Figure~\ref{fig:hdri_scenes}. In both the cases, we would randomize the material properties, lighting intensity and color, as well as adding post-processing effects. Full details on rendering randomizations are provided in Table~\ref{tab:rand_params}. 

In addition to rendering randomizations, we also apply data augmentations during training (with a probability of 80\%) to increase the variety of the training set. A full list of data augmentations and their probabilities are provided in Table~\ref{tab:augmentations}.  Some basic data augmentations are changing the images contrast, brightness, and colour enhancement values. This was done to mimic changes in lighting conditions without having to render a new image. Random background augmentations use the ground truth segmentation to paste a random background on the areas of the image that are not occupied by objects. This was done to avoid overfitting to the 50 HDRI images. We used background images from the VOC dataset~\cite{Everingham10}. Random blend involves randomly alpha blending the image with one of the background images with alpha values chosen uniformly between 0.05 and 0.12. The rationale behind this was to use this to simulate reflections on objects. JPEG compression was also employed. Shot noise was added to make the model robust to low quality images. The snow augmentation adds random white splats that resemble snow in an attempt to simulate specular highlights. The reflectance augmentation originates from intrinsic image decomposition which states that an image, $I$, can be decomposed into a reflectance component, $R$, and a shading component, $S$, such that $I = R \cdot S$. This reflectance component represents the albedo colour of the material whereas the shading component represents the component that is produced by the interaction of lighting and geometry~\cite{garces2021survey}. In this augmentation, to further increase the material diversity, we multiply our original image with a random reflectance map taken from~\cite{BigTimeLi18}. Histogram equalization is the process of evening the distribution of colour intensities and is another augmentation that acts as a way to change lighting conditions. The random perspective data augmentation randomly applies a homography to the image to simulate different camera models. Lastly, we also employ large-scale jittering~\cite{ghiasi2021simple} to resize images, and PASTA~\cite{chattopadhyay2023pasta} which changes the variance in the high frequency components of the image to more closely match the variance of the amplitude spectra of real images.

\begin{figure*}[h]
\centering
    \includegraphics[width=0.4\textwidth]{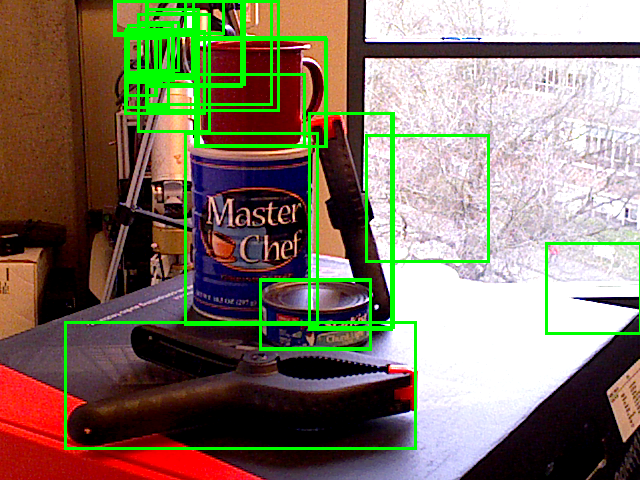}
    \hfill
    \includegraphics[width=0.4\textwidth]{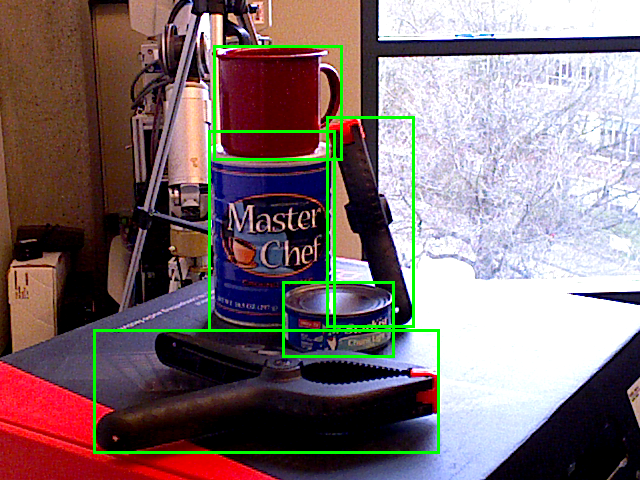}
\caption{\small{Illustration of the importance of confidence thresholds. \textit{Left}: The bounding box predictions of GDet2023 used to get their SOTA results as a result of no confidence threshold. \textit{Right}: The bounding box predictions of our model with confidence threshold of 0.9 which outperforms GDet2023.}}
\label{fig:conf_example}
\end{figure*}

\begin{figure}[h]
\centerline{
\includegraphics[width=0.8\linewidth]
{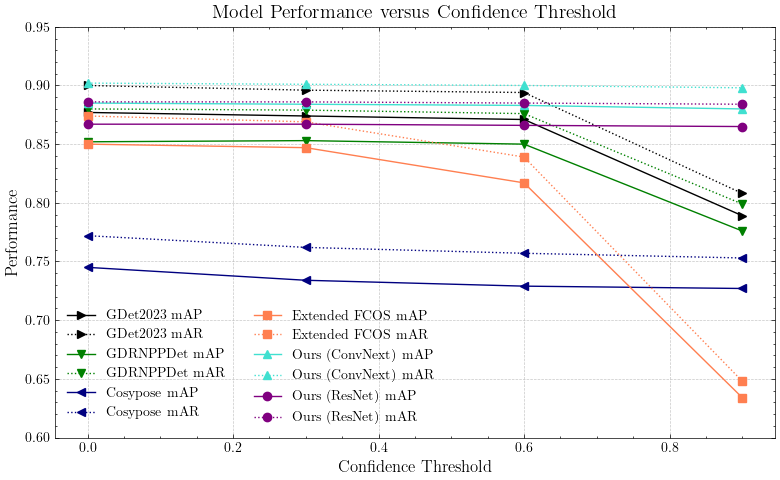}
}
\caption{The mAP and mARs of the best models trained on synthetic+real data as a function of confidence threshold. Predictions with a confidence lower than the threshold are not considered.} \vspace{1mm}
\label{fig:confidence_thresholds}
\end{figure}

\begin{figure}[h]
\centerline{
\includegraphics[width=1.05\linewidth]
{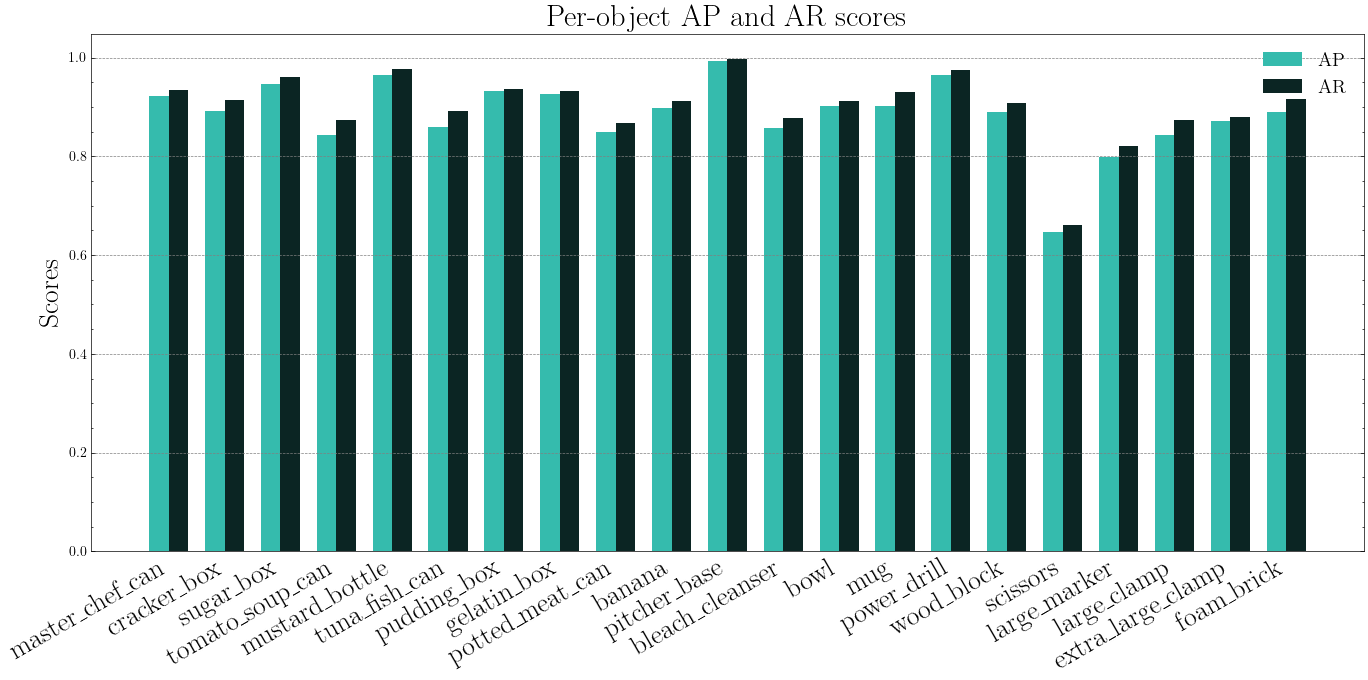}
}
\caption{The AP and AR scores of our best model (ConvNext-S backbone trained on synthetic and real data) for individual objects in the YCBV dataset.} \vspace{1mm}
\label{fig:per_object_scores}
\end{figure}

\begin{figure}[!htp]
\begin{subfigure}{0.5\textwidth}
  \centering
  \includegraphics[width=\linewidth]{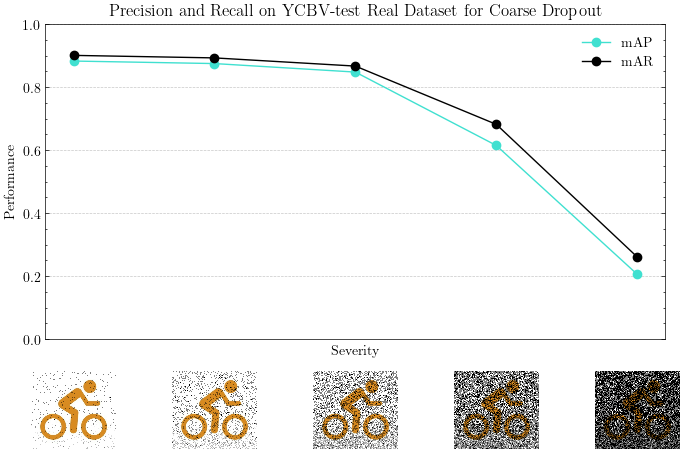}
\end{subfigure}
\begin{subfigure}{0.5\textwidth}
  \centering
  \includegraphics[width=\linewidth]{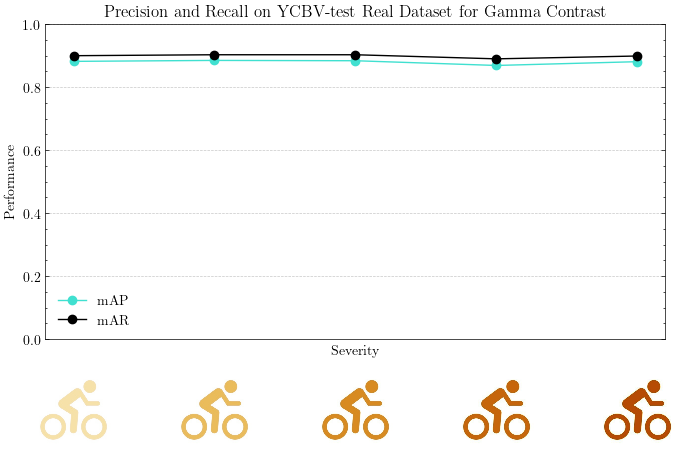}
\end{subfigure}
\vspace{2pt}
\begin{subfigure}{0.5\textwidth}
  \centering
  \includegraphics[width=\linewidth]{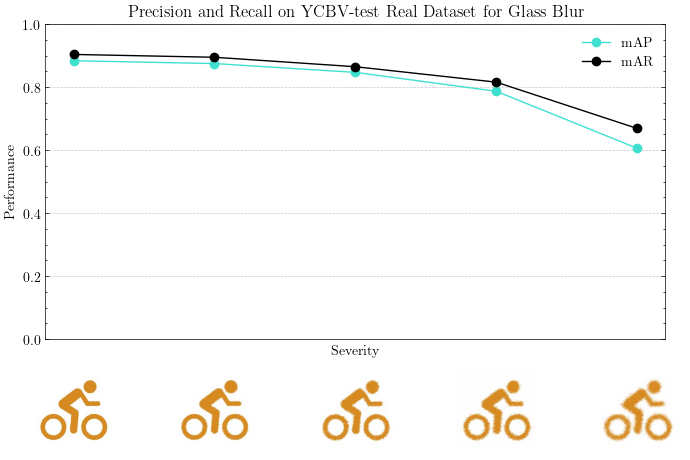}
\end{subfigure}
\begin{subfigure}{0.5\textwidth}
  \centering
  \includegraphics[width=\linewidth]{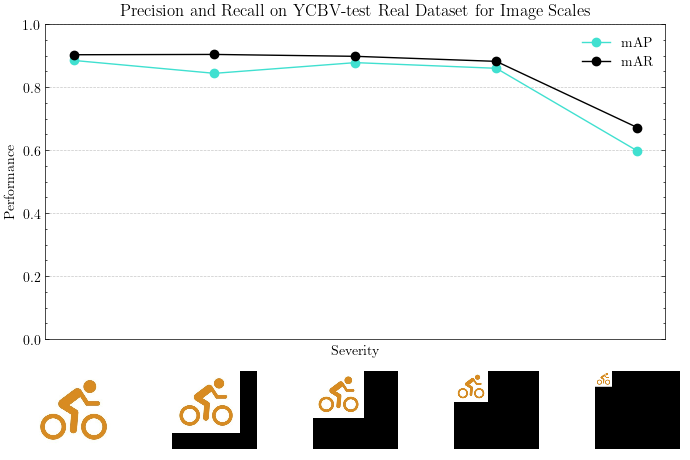}
\end{subfigure}
\vspace{2pt}
\begin{subfigure}{0.5\textwidth}
  \centering
  \includegraphics[width=\linewidth]{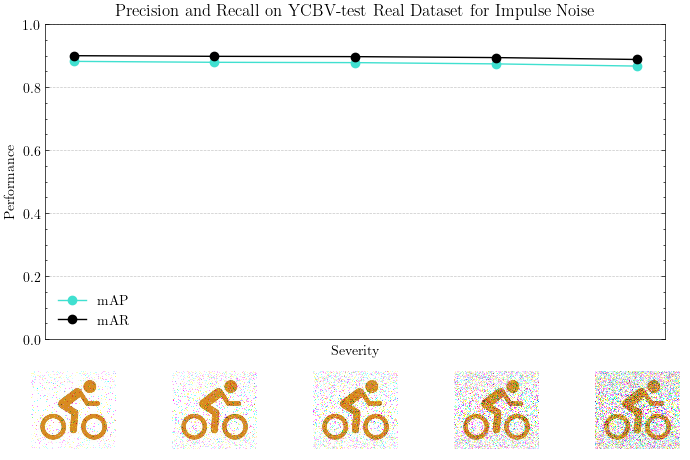}
\end{subfigure}
\begin{subfigure}{0.5\textwidth}
  \centering
  \includegraphics[width=\linewidth]{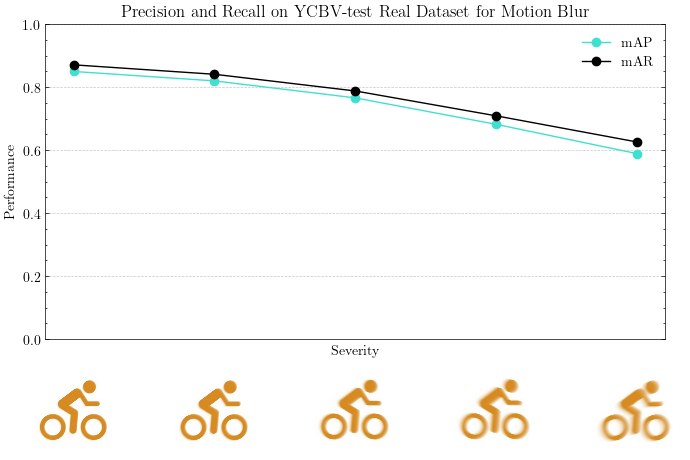}
\end{subfigure}
\caption{Plots characterising the robustness of our best model on different data augmentations. Augmentations such as gamma contrast, glass blur, impulse noise, and motion blur were selected to reflect common corruptions/variations of image capture in the real world. Coarse dropout was chosen to reflect different degrees of occlusion on the objects. The image scales are meant to demonstrate the robustness to smaller resolutions.}
\label{fig:ablations}
\end{figure}

\begin{table}[h]
\centering
\caption{Randomization Parameters}
\label{tab:rand_params}
\begin{tabular}{lll}
\toprule
\multicolumn{3}{c}{Material Properties (once every 20 frames)} \\
\midrule
Parameter    & Distribution & Range \\
\midrule
Albedo Desaturation  & Uniform      & (0.0, 0.4) \\
Albedo Add  & Uniform      & (-0.03, 0.5) \\
Albedo Brightness  & Uniform      & (3.0, 4.0) \\
Diffuse Tint  & Uniform      & ((0.2, 0.2, 0.2), (1, 1, 1)) \\
Reflection Roughness Constant  & Uniform      & (0.5, .7) \\
Metallic Constant  & Uniform      & (0.5, 0.55) \\
Specular Level  & Uniform      & (0.0, 1.0)\\
Emissive Color  & Uniform      & ((0.0, 0.0, 0.0), (0.3, 0.3, 0.3)) \\
\midrule
\multicolumn{3}{c}{Post-Processing (once per frame)} \\
\midrule
Enable TV Noise  & Bernoulli      & (True: 0.1, False: 0.9) \\
\hspace{5mm} Enable Scan Lines  & Bernoulli      & (True: 0.1, False: 0.9) \\
\hspace{10mm} Scan Line Spread  & Uniform      & (0.1, 0.2) \\
\hspace{5mm} Enable Vertical Lines  & Bernoulli      & (True: 0.1, False: 0.9) \\
\hspace{5mm} Enable Random Splotches  & Bernoulli      & (True: 0.1, False: 0.9) \\
\hspace{5mm} Enable Film Grain  & Bernoulli      & (True: 0.1, False: 0.9) \\
\hspace{10mm} Grain Amount  & Uniform      & (0.0, 0.1) \\
\hspace{10mm} Grain Size  & Uniform      & (0.7, 1.0) \\
\hspace{10mm} Color Amount  & Uniform      & (0.0, 0.15) \\
\hspace{5mm} Enable Vignetting  & Bernoulli      & (True: 0.1, False: 0.9) \\
\midrule
\multicolumn{3}{c}{Lighting} \\
\midrule
Ambient Light Intensity (every frame) & Uniform      & (0.1, 0.5) \\
HDRI Background  (every 2K frames) & Uniform      & A set of 50 HDRI backgrounds \\
\midrule
\multicolumn{3}{c}{Configuration} \\
\midrule
Object Height  & Uniform      & (1.0, 5.0) \\
Camera Position  & Uniform      & Sphere of radius 1.0\\
\bottomrule
\end{tabular}
\end{table}

\begin{table}[ht]
\centering
\caption[Data Augmentations]{Data Augmentations and Their Probabilities}
\label{tab:augmentations}
\begin{tabular}{@{}lc@{}}
\toprule
Data Augmentation & Probability \\ \midrule
Colour Contrast       & 0.25         \\
Colour Brightness       & 0.25         \\
Colour Enhancement       & 0.25         \\
Random Background       & 0.60         \\
Random Blend           & 0.40         \\
JPEG Compression          & 0.30         \\
Shot Noise           & 0.40         \\
Snow          & 0.30         \\
Reflectance          & 0.20         \\
Histogram Equalization & 0.40         \\
Random Perspective & 0.45         \\
PASTA & 0.30         \\
Large Scale Jittering & 0.40         \\\bottomrule
\end{tabular}
\end{table}

\subsection{Network Details}
We trained an RT-DETR~\cite{lv2023detrs} model and experimented with both a ResNet-50 and ConvNext-S backbone. All training images were 640$\times$480; however, because the model only accepts 640$\times$640 inputs, we pad the rest of the tensor with zeros. For every image, the model outputs a fixed number of 300 bounding box predictions, most of which can be pruned with confidence thresholding (more details in Section~\ref{sec:results}). The models were trained with the AdamW optimizer~\cite{loshchilov2019decoupled} on an H100 GPU for 20 epochs.


\section{Results}\label{sec:results}
In this section, we demonstrate our results on the YCBV dataset. We also perform ablations on different confidence thresholds and evaluate the robustness of the network to various augmentations that are common in robotics settings, similar to~\cite{hendrycks2019robustness}. Lastly, we demonstrate the ways in which our pipeline can be implemented in the real world.

\subsection{YCBV Benchmark}
\textbf{SOTA Benchmarking Results.} We follow the same method as the official BOP 2022 Challenge for the YCBV dataset~\cite{sundermeyer2023bop} when evaluating object detectors. More specifically, we evaluate our detectors using the mean average precision (mAP) and mean average recall (mAR) metrics. We performed two experiments: one where we only trained our network on synthetic data, and another where we trained on both synthetic and real data. We also experimented with using a detection transformer with both a ResNet-50~\cite{he2015deep} backbone and a ConvNext-S~\cite{liu2022convnet} backbone. The results of these experiments and how they compare with prior works are summarized in Tables~\ref{tab:sim_results} and~\ref{tab:sim_real_results}. Both of these networks were far faster than any of the other networks, with the fastest being the one with the ResNet-50 backbone as it is smaller than ConvNext-S. However, we achieved the best results with the ConvNext-S backbone. The ResNet-50 backbone was 1\% worse in the case of only using synthetic data and 2\% worse when using synthetic and real data. Nonetheless, it is only 1.1\% worse than the prior state-of-the-art while being an order of magnitude faster. All our methods run at real-time speeds, making them useful in all robotics applications and crucial for highly dynamic scenarios. In contrast, the best prior method, GDet2023-PBR, is too slow to use in real-time applications such as robotics as it takes almost 0.2 seconds to compute detections for one image.

All our benchmarking was done with fp32 precision. Using fp16, we can get average inference speeds of 0.01s (91Hz) for the ConvNext-S backbone and 0.004s (230Hz) for the ResNet-50 backbone, with only a minor drop in performance.

\begin{table}[h!]
\centering
\caption[Sim Results]{Sim-only Results}
\label{tab:sim_results}
\begin{tabular}{@{}lc@{\hspace{10pt}}lc@{}}
\toprule
Method & mAP$\uparrow$ & mAR$\uparrow$ & Time (s)$\downarrow$  \\ \midrule
CosyPose-ECCV20-PBR-1VIEW & 0.594 & 0.655 & 0.046 \\
Extended FCOS-PBR & 0.735 & 0.785  &  - \\ 
DLZDet-PBR1  & 0.770 & 0.810 &  - \\ 
GDRNPPDet\_PBR  & 0.786 & 0.830 & 0.079 \\ 
GDet2023-PBR & 0.823 & 0.857 & 0.196 \\ 
Ours w/ ResNet-50 Backbone & 0.821 & 0.844 & \textbf{0.010} \\
Ours w/ ConvNext-S Backbone & \textbf{0.839} & \textbf{0.860} & 0.020 \\
\bottomrule
\end{tabular}
\end{table}

\begin{table}[h!]
\centering
\caption[Sim+Real Results]{Sim+Real Results}
\label{tab:sim_real_results}
\begin{tabular}{@{}lc@{\hspace{10pt}}lc@{}}
\toprule
Method & mAP$\uparrow$ & mAR$\uparrow$ & Time (s)$\downarrow$  \\ \midrule
CosyPose-ECCV20-SYNT+REAL-1VIEW & 0.745 & 0.772 & 0.042 \\
Extended FCOS-MixPBR & 0.850 & 0.874  &  0.030 \\ 
GDRNPPDet\_PBRReal  & 0.852 & 0.880 & 0.082 \\ 
GDet2023 & 0.877 & 0.900 & 0.188 \\ 
Ours w/ ResNet-50 Backbone & 0.866 & 0.884 & \textbf{0.010} \\
Ours w/ ConvNext-S Backbone & \textbf{0.885} & \textbf{0.903} & 0.020 \\
\bottomrule
\end{tabular}
\end{table}

\textbf{Varying Confidence Thresholds.} All of the benchmarking results, including ours, were performed with no confidence threshold as it empirically maximizes the mAP and mAR scores. However, when using a detector in a robotics setting, it is not practical to have no confidence threshold. Instead, it is common to only consider detections with a high confidence value. Figure~\ref{fig:conf_example} illustrates the significance of confidence thresholds. On the left are the predicted bounding boxes from the the prior state-of-the-art, GDet2023, with no confidence threshold. On the right shows the predicted bounding boxes from our methods with a confidence threshold of 0.9. It is clear from here that there are a large number of extraneous bounding boxes when there is no confidence threshold, making it impractical for downstream applications. However, for most models, once a confidence threshold is applied, the performance is significantly limited. Figure~\ref{fig:confidence_thresholds} shows how the different models (trained on both synthetic and real data) perform on the YCBV test set when their confidence thresholds are varied. Most models display a sudden drop in performance when the threshold is set to a high number like 0.9. The only models that are robust to these shifts are the CosyPose detector and ours; although it is important to note that there is a visible decrease in performance for CosyPose as well, not to mention its nominal performance is more than 10\% lower than ours. Overall, this indicates that most models are poorly calibrated and do not have very confident predictions as they tend to label true positives with low confidence. In contrast, our trained models are robust across all confidence thresholds. At 0.9 confidence threshold, we are able to achieve a mAP of 0.880 (still state-of-the-art) and mAR of 0.898. This further highlights the superiority of our models in practical robotics applications.

\textbf{Per-object Performance Breakdown.} Figure~\ref{fig:per_object_scores} shows the breakdown of AP and AR scores across all the objects of the YCBV dataset for our best model. It performs best on relatively larger objects such as the pitcher base, mustard bottle, and power drill. In contrast, we see performance significantly deteriorates for smaller objects such as scissors or large marker. The objects where it performs the worst have very small dimensions along one or more axes which makes detection incredibly difficult if the camera is looking at them from an oblique view. 

\subsection{T-LESS Benchmark}
\begin{figure*}[ht]
    \includegraphics[width=\textwidth]{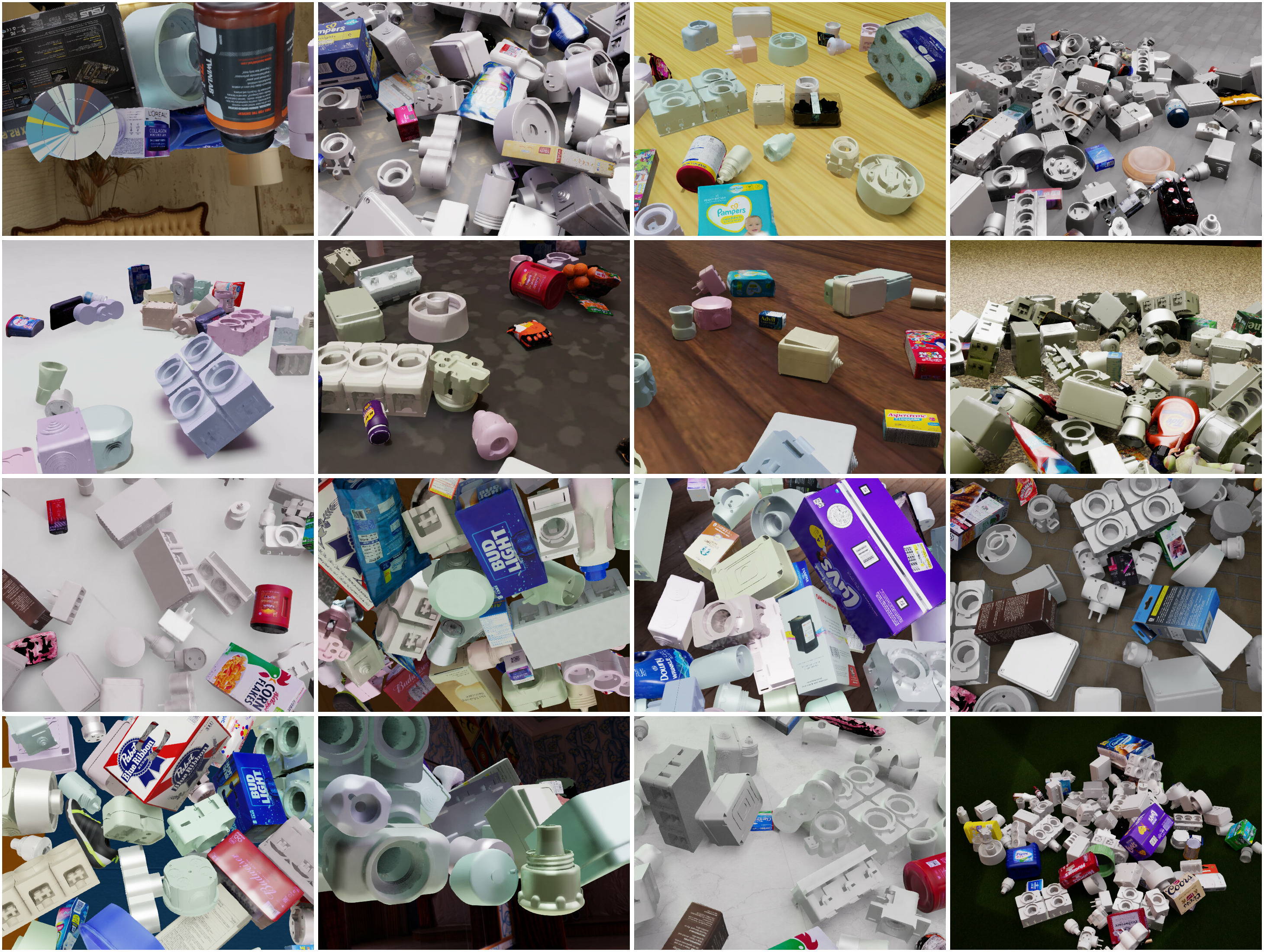}
    \hfill
\caption{Example images generated for T-LESS dataset.}
\label{fig:tless_montage}
\end{figure*}

In order to demonstrate the applicability of this method for other object datasets, we also benchmarked on the T-LESS dataset which contains a wide variety of textureless, industrial objects. This presents a distribution of objects that is substantially different from that of the YCBV dataset. Figure~\ref{fig:tless_montage} contains a montage showing the various images generated for the T-LESS dataset. Using the ConvNext-S backbone, we were able to achieve a mAP of 0.890 which is 0.003 less than the state-of-the-art, and achieved the best mAR of 0.817. Contrary to the YCBV benchmarking, for T-LESS, most methods (including ours) had relatively higher mAP scores and lower mAR scores. This implies that most models struggle to regress true positives. We hypothesize that this was because the texture-less nature of the objects makes them harder to discern from the background.

\begin{table}[h!]
\centering
\caption[Sim+Real Results]{Sim+Real Results}
\label{tab:sim_real_results}
\begin{tabular}{@{}lc@{\hspace{10pt}}lc@{}}
\toprule
Method & mAP$\uparrow$ & mAR$\uparrow$ & Time (s)$\downarrow$  \\ \midrule
CosyPose-ECCV20-SYNT+REAL-1VIEW & 0.693 & 0.681 & 0.042 \\
Extended FCOS-MixPBR & 0.798 & 0.751  &  0.030 \\ 
GDRNPPDet\_PBRReal  & 0.876 & 0.801 & 0.082 \\ 
GDet2023 & \textbf{0.894} & 0.815 & 0.188 \\ 
Ours w/ ConvNext-S Backbone & 0.890 & \textbf{0.817} & 0.020 \\
\bottomrule
\end{tabular}
\end{table}

\subsection{Robustness Experiments}
In the real world, corruptions or simple variations in the image intensities can occur during the image acquisition process. For models being deployed in safety-critical applications such as robotics, it is important for practitioners to study and understand their failure cases. In Figure~\ref{fig:ablations}, we benchmark our best model across varying severities of image corruptions. We use image corruptions such as gamma contrast, glass blur, impulse noise, and motion blur, similar to ~\cite{hendrycks2019robustness}. We further characterise the robustness by also including varying image scales and coarse dropout intensities.

For gamma contrast and impulse noise, the model is robust across the different levels of severities. This is probably because as shown in Table~\ref{tab:augmentations}, we include camera shot noise and colour contrast variations in our training-time data augmentations. Our model is fairly robust at different image resolutions until the image dimensions are scaled a lot to 20\% of the original value which is when model performance severely degrades. The model is not robust to extreme cases of coarse dropout, glass blur, or motion blur, suggesting that they should be included in the training time data augmentations for increased robustness.

\subsection{Real World Experiments}
\begin{figure}[h]
\centerline{
\includegraphics[width=1\linewidth]
{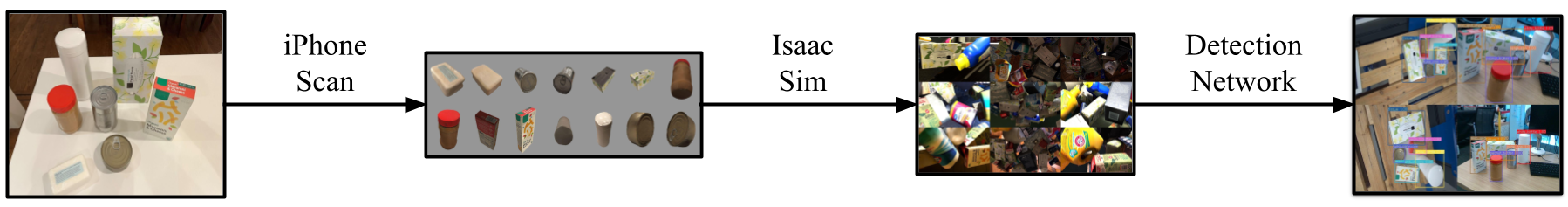}
}
\caption{The pipeline we use to create custom datasets. We first use the AR Core app available for the iPhone Pro to get 3D scans of objects. The objects are then converted to USD and imported into Omniverse Isaac Sim in order to perform data generation. This can then be used to train object detectors for use in the real world.} \vspace{1mm}
\label{fig:scan_workflow}
\end{figure}

In order to demonstrate the practicality of our pipeline in a real-world setting, we showcase how detectors can be trained for arbitrary objects. The pipeline is shown in Figure~\ref{fig:scan_workflow}. First, a complete scan of an object can be made using the AR Code Object Capture 3D Scan app, available on all iPhone Pro or iPad Pro models with a LiDAR sensor. It can take anywhere between 5 and 10 minutes to scan a single object. The asset can then be converted into USD format and brought into Isaac Sim for data generation similar to how data was generated for the YCB objects in Section~\ref{sec:approach}. Even though the asset model for these objects is fairly low quality as it was generated by an iPhone, as opposed to the laser scanner used for YCB objects~\cite{xiang2018posecnn}, it still suffices for generating data that allows the model to generalise to the real world. 

\subsection{Limitations}
While we are able to improve upon the state-of-the-art in terms of both performance and speed, there are still some important limitations to our work. First, the data generation is restricted to objects for which we have CAD models. As a result, it does not work on unseen objects. This presents a promising research direction for scaling synthetic data for open-vocabulary detectors such as LLava~\cite{liu2023llava,liu2023improvedllava} and GLEE~\cite{wu2023GLEE}. Secondly, even though our detector has the best performance on the benchmarks, when used in the real world it is still prone to occasional missed detections if there is significant motion blur and occlusion. As seen in Figure~\ref{fig:per_object_scores}, it is clear that model performance heavily varies depending on the object. Further research could investigate data generation methods that emphasize the presence of smaller objects to ensure parity across all the objects in the dataset. Lastly, the generated data was IID and was agnostic of the model. Future work could investigate areas of active learning to bring model learning in-the-loop of data generation in order to increase the amount of images generated that are useful for learning and reducing the amount of redundant information in the dataset.

\section{Discussion}
\subsection{Implementation Tips}
\begin{figure}[ht]
\centering

\begin{subfigure}[t]{0.48\textwidth} 
    \centering
    \includegraphics[width=\linewidth,height=2.25in,keepaspectratio]{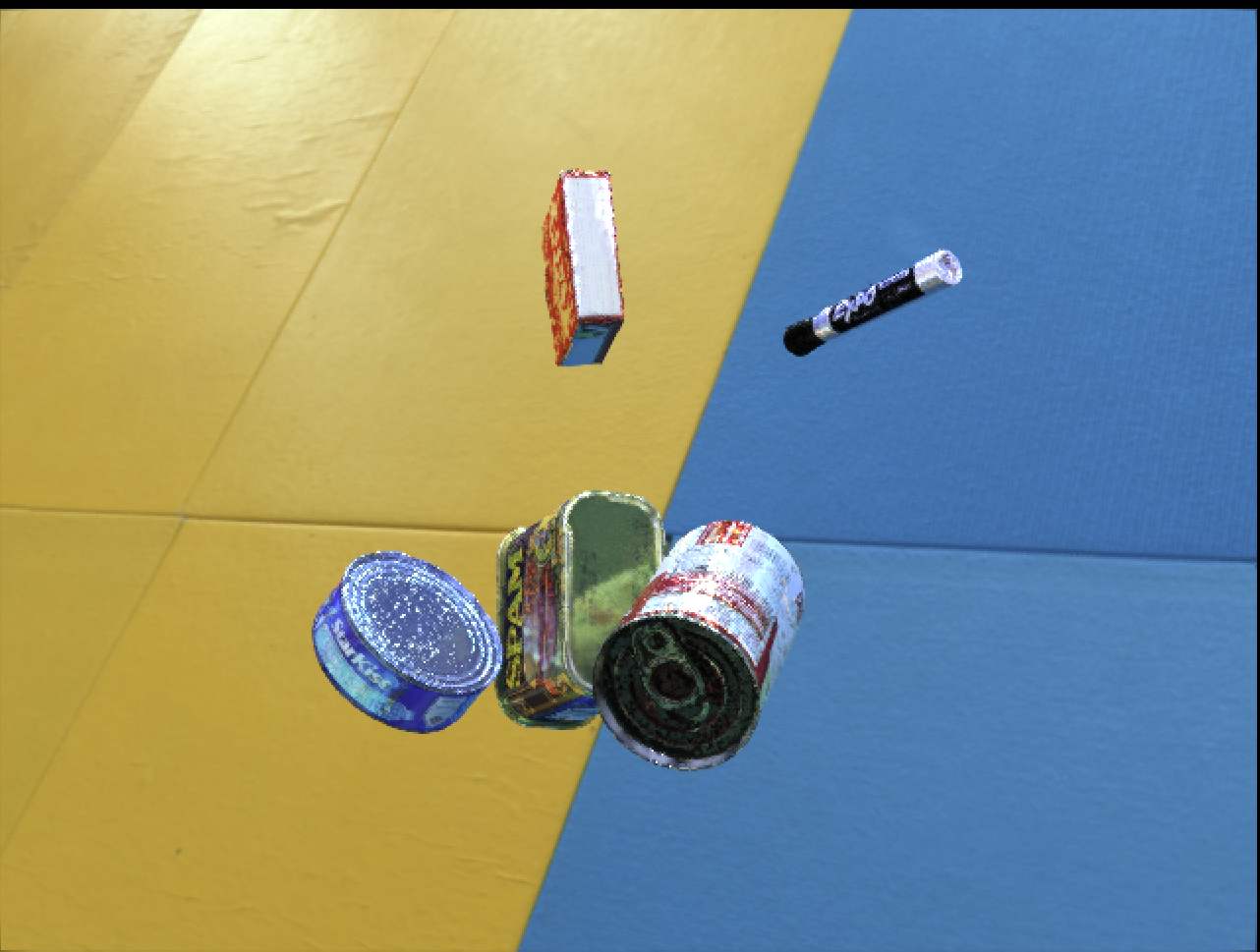} 
    \label{fig:denoise_imagess_sub1}
\end{subfigure}
\hfill 
\begin{subfigure}[t]{0.48\textwidth} 
    \centering
    \includegraphics[width=\linewidth,height=2.25in,keepaspectratio]{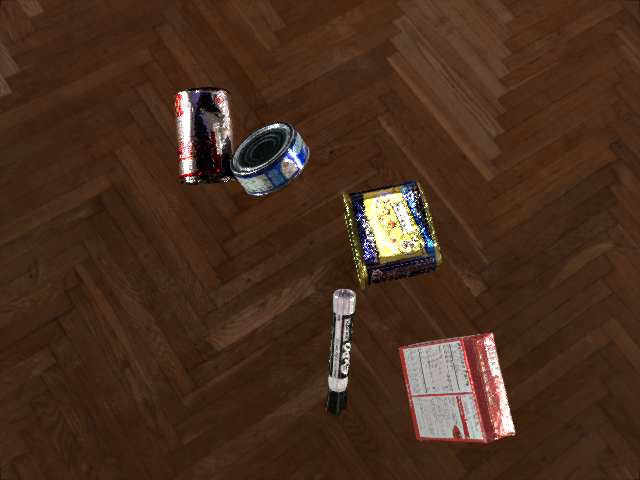} 
    \label{fig:denoise_images_sub2}
\end{subfigure}

\caption[Denoiser Artifacts]{Sample images of denoiser not receiving enough samples.}
\label{fig:denoise_images}
\end{figure}

With this, there were some key lessons learnt for data generation. Firstly, setting a very low ambient light intensity typically leads to more realistic looking images. This intensity is not caused by any light sources, and in the real world, there is roughly 0 ambient light intensity as all illumination is either directly or indirectly caused by a light source. By setting it to be very low, it allows a larger portion of the intensity of an object to come from secondary ray bounces thereby making the indirect lighting more prominent, hence enhancing the realism. However, we randomized this and allowed it to be as high as 0.5 because low ambient light tended to lead to far darker images that required a lot of tuning of other light sources. Also, having darker scenes in general makes it harder for the denoiser to output a clean image (as it needs to sample far more light rays) which leads to rendering artifacts as shown in figure~\ref{fig:denoise_images}. The tuna can in the left image and the tomato soup can in the right image show some noise or dark patches caused by the denoiser not being able to adequately render them. Additionally, most assets do not have authored materials and simply just give an albedo texture. As such, it can make apparent metallic-like objects (e.g. potted meat can, tuna can, etc.) behave more like a piece of paper. In order to mitigate this, the metallic constant can be set to 1. It typically is meant to be either a 0 or 1 as whether a material is a metal or non-metal affects the colour of the specular reflection~\cite{pharr2016physically}. We noticed that setting the metallic value to something in between can also give acceptable looking images. In order to make materials behave more like metals, the roughness parameter could also be reduced which makes the material smoother thereby making the reflections more noticeable. However, when decreasing roughness, this causes issues with denoising artifacts that are also seen with darker images. This is because the denoiser needs more samples from secondary rays to accurately reconstruct the reflection. A solution that does not require adding too many extra light sources is to increase the subframe count. Subframes essentially accumulate light transport over several frames to get a more accurate image with less noise. This should be increased when there are more objects in the scene and/or when lowering roughness so that the denoiser gets more samples. The only downside to this is that it takes more time to render each individual frame. For more realistic and soft shadows, one method is to use larger light sources (with a correspondingly lower intensity to not overexpose the object). These will make the rendering slower but give a soft transition into the shadow which is typical of the real world. If speed is desired, point lights can be used which are very efficient as illumination of a fragment does not require any sampling, but it comes at the downside of creating very sharp boundaries between shadow and non-shadow regions. Lastly, if there is a randomization that is randomly making objects visible or invisible, instead of randomly changing the visibility setting on the object, it is faster to simply move the object out of frame (e.g. simply setting the z-value to a very large number). This is because setting an object to invisible destroys the asset in the rendering backend and it becomes costly to create the asset again whereas simply moving it out of the view frustum is far more efficient.

\subsection{Test-Time Augmentations}
Some methods also employ the use of test-time augmentations. An example of one such method is the previous state-of-the-art network, GDET2023~\cite{wang2021gdrnet}. This is a technique where when benchmarking a model on an image, the image is augmented in several different ways and the model outputs predictions on all of these different augmented images and uses some method to then aggregate all these predictions. It is akin to essentially ensembling different model predictions in order to make the outputs more robust. They find that it can improve their benchmarking results. For example, with their best performing checkpoint, they achieve their best results with mAP of 0.877 and mAR of 0.900 while employing test-time augmentations. Benchmarking their best checkpoint without these test-time augmentations, we find it reduces their performance to a mAP score of 0.851 and mAR score of 0.880. The issue with these test-time augmentations is that they make object detection slower as now the image needs to be transformed in various different ways. This is likely why their inference time on the leaderboards is far slower than the inference time of their network, as a lot of time is probably spent in data-preprocessing for augmenting the images. In contrast, our state of the art results with mAP of 0.885 and mAR of 0.903 were achieved with no test-time augmentations. This means that the true performance gap between our method and the previous state of the art is larger than it originally seems. While we could also try to incorporate test-time augmentations with our method (and this would likely push our numbers up even more), it is likely that it would just overfit. These augmentations are typically tuned by testing which one works the best on the test set, but that would cause the entire detection pipeline to overfit to the test distribution and ultimately reduce the robustness of the detection method in novel scenarios. 

\subsection{Analysing the mAP Metric}
The drop in performance demonstrated in Figure~\ref{fig:confidence_thresholds} begs the question of whether the mean average precision is the correct metric of success for detection networks in robotics. The average precision is calculated as the area under the interpolated precision-recall curve and it is justified by claiming that one would want to go through a few extra false positives to arrive at one extra true positive. This may certainly have been an apt trade-off in the metric's original field of application, information retrieval, because a false positive is not as harmful as the user would simply just need to scroll through more links~\cite{manning2008introduction}. In contrast, for a robotics application, a false positive can be very detrimental to safety as a robot may erroneously try interacting with the environment which may cause damage. It is clear that a new metric needs to be used for robotics detection networks that is able to appropriately weigh the safety of the robot and the task completion. 

Precision can be seen as a way of prioritizing robot safety as maximizing precision will minimize the percentage of false positives which will improve the safety of the robot. On the other hand, recall can be seen as prioritizing task completion as maximizing recall will maximize the number of true positives which will give the robot more opportunities to interact with the environment, and thus, complete its task. While strictly prioritizing one metric over the other is insufficient, there must be a way to balance both, with an emphasis on precision to ensure that non-trivial tasks can be completed in a safe manner. While the aim of this work is not to propose any new such metric, one area of possible future research is to potentially investigate methods that leverage an F-score where $\beta < 1$. When $\beta < 1$, this means that the F-score weighs precision more than it weighs recall. The famous F1 score weighs them both equally (i.e. $\beta = 1$). 

Giving preferential weighting to precision or recall has already been done in the fields of information retrieval and recommenders. For example, when looking at patent search in information retrieval, practitioners typically try to optimize metrics that prioritize recall over precision. This is to ensure that users could guarantee with a high probability that when they conduct a prior art search, they would receive all the relevant prior art~\cite{manning2008introduction}. In contrast, if someone wanted to test a recommender system that returns the top-$k$ recommendations, it would be more beneficial to evaluate with a metric that prioritizes precision to reduce the number of irrelevant recommendations that the user receives~\cite{manning2008introduction}.

\begin{figure}[!htbp]
\centerline{
\includegraphics[width=0.6\linewidth]
{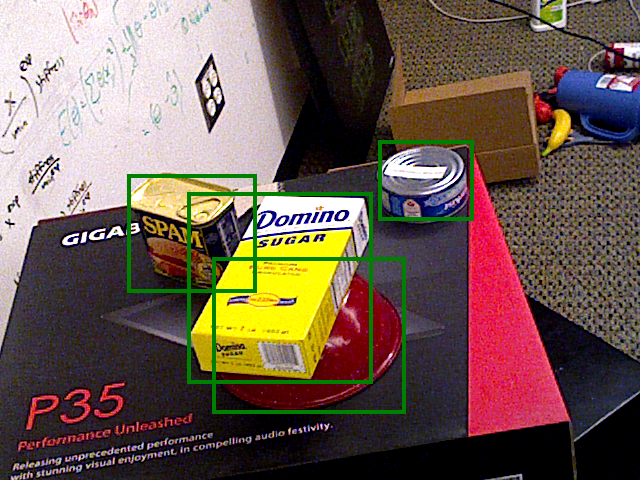}
}
\caption{A data sample taken from the YCB-Video dataset. The groundtruth bounding boxes are overlayed on the sampled image. It is clear that the groundtruth annotations are missing bounding boxes for the bleach cleanser, banana, tomato soup can, pitcher base, and drill located in the top right corner.} \vspace{1mm}
\label{fig:missing_annotations}
\end{figure}

\section{Conclusion}
We present Synthetica, a method for generating large-scale, synthetic images for robotics tasks. Our main contribution is providing an object detector that achieves state-of-the-art results on existing detection benchmarks while also having the fastest inference time. Our approach demonstrates the utility of scaling up synthetic data generation for training robust vision networks. This is in contrast with other methods which rely on collecting real-world data, which is both time-consuming and difficult to scale. The speed and performance of our detection networks ensure that they can be a practical component in existing robotics control stacks that have very strict timing requirements. Lastly, we demonstrate a simple pipeline for creating  custom object detectors for which there are no prior real-world datasets.

\clearpage
%
%
\bibliographystyle{splncs04}
\bibliography{egbib}
\end{document}